\documentclass[11pt]{article}

\usepackage[utf8]{inputenc}
\usepackage[margin=1in]{geometry}
\usepackage{amsmath,amssymb}
\usepackage{graphicx}
\usepackage{booktabs}  
\usepackage{algorithm}
\usepackage{algpseudocode}  
\usepackage{listings}  
\usepackage{xcolor}
\usepackage{hyperref}  
\usepackage{pgfplots}  
\usepackage{tikz}      
\usepackage{float}     
\usepackage{textcomp}
\usepackage{ifthen}    
\usetikzlibrary{shapes,arrows,positioning,calc}
\pgfplotsset{compat=1.18}

\graphicspath{{./}}

\newcommand{\includefig}[2][]{%
  \IfFileExists{#2}{\includegraphics[#1]{#2}}{\centering\fbox{\parbox{0.85\linewidth}{\centering\small [Figure: #2]}}}%
}

\hypersetup{
    colorlinks=true,
    linkcolor=blue,
    citecolor=blue,
    urlcolor=blue
}

\lstdefinestyle{cuda}{
    language=C++,
    basicstyle=\ttfamily\small,
    keywordstyle=\color{blue}\bfseries,
    commentstyle=\color{gray}\itshape,
    stringstyle=\color{red},
    numbers=left,
    numberstyle=\tiny\color{gray},
    stepnumber=1,
    numbersep=5pt,
    backgroundcolor=\color{white},
    showspaces=false,
    showstringspaces=false,
    showtabs=false,
    frame=single,
    tabsize=2,
    captionpos=b,
    breaklines=true,
    breakatwhitespace=false,
    escapeinside={(*@}{@*)},
    morekeywords={__global__, __device__, __host__, __shared__, __syncthreads}
}

\title{\textbf{GPUTOK: GPU Accelerated Byte Level BPE Tokenization}}

\author{
    Venu Gopal Kadamba \\
    M.S.\ in Data Science \\
    New York University \\
    \texttt{vk2636@nyu.edu}
    \and
    Kanishkha Jaisankar \\
    M.S.\ in Data Science \\
    New York University \\
    \texttt{kj2675@nyu.edu}
}

\date{}  

\begin{document}

\maketitle

\begin{abstract}
As large language models move toward million-token context windows, CPU tokenizers become a major slowdown because they process text one step at a time while powerful GPUs sit unused. We built a GPU-based byte-level BPE tokenizer that follows GPT-2's merge rules. It includes a basic BlockBPE-style kernel and a faster, optimized version that uses cuCollections \texttt{static\_map}, CUB reductions, and a \texttt{pybind11} interface for Python. 

On WikiText103 sequences up to 131k tokens, the optimized GPU tokenizer produces the same tokens as a CPU version and, for the longest inputs, is about \(1.7\times\) faster than \texttt{tiktoken} and about \(7.6\times\) faster than the HuggingFace GPT-2 tokenizer. Nsight profiling shows that 70--80\% of CUDA API time goes to memory allocation, so adding memory pooling should give the biggest speed boost next. Tests on generation tasks using WikiText103 prompts show that our GPU tokenizer's outputs stay within about one percentage point of \texttt{tiktoken} and HuggingFace GPT-2 on similarity and overlap metrics, meaning it keeps output quality while making long-context inference more practical.
\end{abstract}

\section{Introduction}

Large language models are moving from a few thousand tokens to hundreds of thousands and even million-token context windows. In this long-context regime, it is not enough to only speed up the transformer layers: preprocessing stages such as tokenization can become a major part of end-to-end latency and delay responses even when a powerful GPU is available. If tokenization stays on a single CPU core, the GPU may sit idle while the CPU slowly turns raw text into tokens, which wastes hardware, increases serving cost, and limits real-time applications that need to handle long chats, whole documents, or collections of documents. Even well-tuned CPU tokenizers like \texttt{tiktoken} and HuggingFace tokenizers are mostly sequential and do not exploit the massive parallelism of modern GPUs, so a GPU-aware tokenizer becomes an important piece for keeping latency low as context windows grow.

Byte Pair Encoding (BPE)~\cite{You2025BlockBPE} is widely used in decoder-only LLMs including GPT2~\cite{radford2019gpt2}, but its standard greedy algorithm is iterative and data dependent: it repeatedly finds the best adjacent pair according to a merge table and merges its left-most occurrence until no more merges are possible. This behavior is hard to map efficiently to GPUs because each merge pass depends on the previous one, the sequence length changes after every merge, and real-world vocabularies require many hash table lookups while still matching a trusted CPU implementation exactly. A design that simply ports the CPU algorithm to the GPU will not perform well; instead, we aim to exploit data parallelism within each merge pass while preserving GPT2-compatible greedy semantics, and we build on the BlockBPE framework~\cite{You2025BlockBPE} to guide this design.

\subsection{Our approach}

We build on the BlockBPE framework~\cite{You2025BlockBPE} to design a GPU implementation of byte-level BPE that is compatible with GPT2-style tokenization and tuned for long-context workloads. At a high level, our approach is:

\begin{enumerate}
    \item \textbf{GPU kernels for GPT2 merges}: We implement two CUDA kernels that follow GPT2 merge priorities: a baseline kernel that mirrors BlockBPE's global-argmin greedy merge, and an optimized kernel that merges non-overlapping local minima per pass to reduce the number of iterations while keeping the same merge order.
    \item \textbf{Simplified compaction}: We use a simple double-buffered compaction scheme that works for all sequence lengths, reduces shared memory usage, and avoids complex prefix-sum logic, making the kernels easier to tune.
    \item \textbf{Python integration}: We wrap the implementation in a pybind11-based Python interface that reconstructs the GPT2 byte encoder, builds a GPU-resident \texttt{static\_map} for merges, and exposes batch tokenization APIs suitable for PyTorch-based serving pipelines.
\end{enumerate}

Section~\ref{sec:background} reviews BPE, BlockBPE, and related work on tokenization and GPU hash tables. Section~\ref{sec:approach} describes our system architecture and kernel design in detail. Section~\ref{sec:experiments} presents the experimental setup and benchmarks, and Section~\ref{sec:results} reports empirical results and profiling analysis. Section~\ref{sec:conclusion} summarizes our findings and outlines directions for future work.

An open-source implementation, including CUDA kernels, Python bindings, and benchmarking scripts, is available at \url{https://github.com/venugopalkadamba/gpu-tokenizer}.

\section{Literature Review}
\label{sec:background}

Tokenization for modern LLMs lies at the intersection of subword algorithms, high-performance systems design, and GPU-oriented data structures. Prior work can be grouped into three main categories: (1) algorithmic variants of Byte Pair Encoding (BPE), (2) systems and implementation strategies on CPU versus GPU, and (3) supporting GPU primitives that enable scalable tokenization. This taxonomy highlights not only how merges are defined, but also how they can be executed efficiently in practical serving environments.

\paragraph{Algorithmic Variants of BPE.}
Classical BPE~\cite{Sennrich2016} learns frequent pairwise merges to build subword vocabularies, and GPT-2 adopts a byte-level variant that avoids invalid UTF-8 sequences by mapping bytes to printable Unicode~\cite{radford2019gpt2}. Follow-up work revisits BPE from broader perspectives: fairness-motivated objectives such as parity-aware BPE~\cite{parityaware2024} aim to equalize compression across languages, while stochastic inference methods such as BPE dropout~\cite{Provilkov2020} explore robustness and variability in merge application. These works reveal a wide design space but focus primarily on how merges should be \emph{chosen}, not on how to execute them under the resource and latency constraints of LLM serving pipelines.

\paragraph{Systems and Implementation Strategies.}
From a systems viewpoint, three families of tokenizers dominate. Regex-based CPU tokenizers, including the original GPT-2 and many HuggingFace implementations, use language-aware pre-tokenization before applying BPE. They provide strong text handling but suffer from expensive regex processing and irregular control flow, which complicates GPU acceleration. Optimized CPU libraries such as \texttt{tiktoken} and HuggingFace Tokenizers push throughput closer to hardware limits through cache-aware engineering, SIMD, and low-overhead bindings. These achieve excellent single-node performance but leave GPUs idle during tokenization, creating a mismatch in GPU-centric inference stacks.

The third line of work moves tokenization onto accelerators. BlockBPE~\cite{You2025BlockBPE} shows that BPE can be parallelized by evaluating all adjacent pairs in each pass and using GPU-resident hash tables plus block-level scans. Several variants relax the strict greedy semantics to allow multiple non-overlapping merges per pass, greatly improving GPU utilization at the cost of producing token sequences that may deviate from CPU reference implementations. Other GPU-oriented efforts adopt byte-level simplifications or approximate merging schemes, improving throughput but reducing compatibility with GPT-2-style vocabularies.

\paragraph{GPU Hash Tables and Parallel Primitives.}
Efficient GPU tokenization depends on device-resident data structures and collective operations capable of running at scale. cuCollections offers high-throughput GPU hash maps such as \texttt{static\_map}, which are well suited for static merge tables. Likewise, CUDA C++ Core Libraries provide CUB's block-level scans and reductions, which have become standard tools in GPU data-parallel workloads. While these components offer the raw building blocks for a GPU tokenizer, the literature does not provide a systematic study of how to assemble them into an exact greedy BPE pipeline compatible with GPT-2 merge rules.

\paragraph{Pros and Cons of Existing Approaches.}
CPU regex tokenizers excel at language-aware behavior and guarantee exact compatibility with training-time tokenization, but suffer from limited parallelism, high preprocessing cost, and CPU--GPU data movement overheads. Optimized CPU libraries significantly improve throughput yet remain fundamentally constrained by CPU parallelism, making them insufficient for long-context, GPU-heavy inference. GPU-accelerated BPE variants demonstrate strong speedups and compelling scalability, but many relax greedy merge semantics or adopt simplified pre-tokenization, limiting their suitability as drop-in replacements for GPT-2-style models.

\paragraph{Why Existing Work Is Not Enough.}
Taken together, prior work leaves a gap at the intersection of exact GPT-2 semantics and GPU-centric systems design. Algorithmic papers do not address system-level performance; CPU libraries ensure correctness but underutilize GPU resources; and GPU implementations prioritize throughput over exact equivalence. Moreover, although cuCollections and CUB provide the necessary primitives, no existing work shows how to combine them into a tokenizer that matches GPT-2 merge priorities, exposes a practical Python API, and is evaluated rigorously across realistic long-context workloads.

This work targets that underexplored space by designing a GPU-resident BPE tokenizer using cuCollections hash maps and block-level reductions, preserving GPT-2-style greedy merges while integrating cleanly into Python-based LLM serving pipelines. The result aims to deliver both exact compatibility and high throughput, addressing the shortcomings of prior CPU- and GPU-centric approaches.

\subsection{Positioning relative to BlockBPE and prior GPU tokenizers}

BlockBPE~\cite{You2025BlockBPE} is, to our knowledge, the first end-to-end GPU implementation of BPE tokenization. It removes Regex-based pre-tokenization, uses GPU-resident hash maps and block-level prefix scans, and demonstrates up to roughly \(2\times\) higher throughput than \texttt{tiktoken} and \(2.5\times\) over HuggingFace Tokenizers on high-batch workloads. In exchange, BlockBPE accepts small deviations from the behavior of canonical CPU tokenizers---most notably by eliminating Regex pre-tokenization---and is explicitly optimized for high-throughput batch inference rather than the long single-sequence regime.

Our work is narrower in ambition but different in focus. We treat GPT-2-style byte-level BPE as fixed and aim for a strict drop-in replacement for widely used CPU tokenizers: we reconstruct the original byte encoder, preserve the greedy merge semantics, and require token-level equality against a CPU reference on all our benchmarks. On the systems side, we explore a specific design point that combines cuCollections \texttt{static\_map} and lightweight, double-buffered compaction within a single CUDA block per sequence, and we integrate the result into a minimal Python API suitable for common PyTorch-based serving stacks.

We do not claim algorithmic novelty over BlockBPE, nor do we attempt to match its extensive exploration of batch regimes or scheduling policies. Instead, our contribution should be read as a focused case study of how a strictly compatible GPT-2 tokenizer can be moved to the GPU with a small and explicit set of engineering choices, along with an empirical analysis of when this design is helpful and where it still falls short.

\section{Proposed Approach}
\label{sec:approach}

We now describe the overall architecture, how we reconstruct GPT2 byte encodings, how we store the merge table on the GPU and how the kernels are structured. In keeping with the BlockBPE~\cite{You2025BlockBPE} setting, our design must simultaneously (1) reproduce GPT2's greedy merge semantics exactly, and (2) expose enough fine-grained parallelism to make running the tokenizer on GPUs worthwhile. The discussion below is organized around these two goals: for each component we explain both \emph{what} it does and \emph{why} we structured it that way.

\subsection{System architecture}

At a high level, tokenization proceeds in the following stages:

\begin{enumerate}
    \item \textbf{Byte encoding (CPU).} Input UTF-8 text is converted to raw bytes and then mapped to GPT2-style symbols via a deterministic byte encoder implemented in C++.
    \item \textbf{Chunking and batching (CPU).} The resulting token sequence is optionally split into fixed-size chunks (up to a configured token budget) so that each chunk fits in GPU shared memory and can be assigned to a single CUDA block.
    \item \textbf{Merge passes (GPU).} For each chunk, one CUDA block loads the tokens into shared memory and runs the greedy BPE merge loop entirely on the device, using a GPU-resident hash table for pair lookups and block-level reductions for argmin.
    \item \textbf{Assembly (CPU).} The merged token sequences from all chunks are copied back, concatenated in the original order, and returned as the final token ids.
\end{enumerate}

In our experiments, the sequence lengths of interest are chosen so that each individual sequence can be processed by a single block, which simplifies synchronization and avoids inter-block communication. For profiling on full books we enable host-side chunking to keep the per-block shared-memory footprint bounded. We considered more complex schemes that shard a single sequence across multiple blocks, but those require additional global coordination to enforce GPT2's left-most greedy semantics, whereas the one-block-per-sequence design makes the merge order deterministically local to the block and therefore much easier to reason about and verify.

\subsection{GPT2 byte encoder reconstruction}

GPT2 uses a custom byte-to-Unicode mapping to avoid unknown tokens and invalid UTF-8 sequences. We re-implement this mapping in C++ following the original reference logic. A simplified version is shown in Listing~\ref{lst:byte_encoder}.

\begin{lstlisting}[style=cuda, caption={Byte to symbol encoding (simplified)}, label={lst:byte_encoder}]
struct ByteEncoder {
  std::vector<std::string> byte_to_symbol;  // size 256

  ByteEncoder() {
    // Map printable ranges to themselves (bytes already "safe" as UTF-8)
    for (int c = '!'; c <= '~'; ++c) { /* ... */ }
    for (int c = 0xA1; c <= 0xAC; ++c) { /* ... */ }
    for (int c = 0xAE; c <= 0xFF; ++c) { /* ... */ }

    // Map remaining bytes to shifted codepoints
    int n = 0;
    for (int b = 0; b < 256; ++b) {
      if (/* not already mapped */) {
        byte_to_symbol[b] = codepoint_to_utf8(256 + n++);
      }
    }
  }
};
\end{lstlisting}

On the host, we first apply the encoder to produce a sequence of 256-way byte-level symbols and then map those symbols to integer token ids using the BPE vocabulary (Section~\ref{sec:background}). Subsequent merges operate purely on integer token ids, so all GPU-side work is done in terms of fixed-width integers. Reconstructing the original GPT2 byte encoder rather than inventing a simpler scheme is important for compatibility: it guarantees that our merge table, byte-level representation, and final ids are aligned with existing GPT2 models and tokenizer implementations.

\subsection{Merge table representation}

The GPT2 merge vocabulary is stored in \texttt{merges.txt} with around fifty thousand merge rules. We parse these into two tables that share the same underlying integer token ids:

\begin{itemize}
    \item An \texttt{std::unordered\_map<PairKey, PairInfo>} on the CPU, used for the reference implementation.
    \item A \texttt{cuco::static\_map<PairKey, PairVal>} on the GPU, used by the CUDA kernels.
\end{itemize}

Each adjacent token pair is represented in packed 64-bit form as
\[
\text{key} = (a \ll 32) \,\,|\,\, b, \quad
\text{value} = (new\_token \ll 32) \,\,|\,\, rank,
\]
where $a$ and $b$ are the input tokens, $new\_token$ is the id of the merged token, and $rank$ is the merge-table rank. Packing into 64-bit values allows us to perform lookups using a single key and to store both the rank and the result in one word, which keeps the GPU-side API simple and reduces device memory traffic.

On the GPU we use \texttt{cuco::static\_map} with linear probing and size it at roughly fifty percent load factor. This choice trades a modest increase in memory footprint for much shorter probe sequences and better coalescing when thousands of threads concurrently query the map. The map is constructed once at initialization time (either when the stand-alone binary starts or when the Python extension is imported) and then reused across all tokenization calls, so the cost of building the hash table is amortized over many batches.

\subsection{Baseline kernel}

The baseline kernel in \texttt{gputok\_blockbpe.cu} is a direct per-block translation of Algorithm~1 in BlockBPE. Each CUDA block is responsible for exactly one chunk, and the block repeatedly applies the single best merge (global minimum rank, left-most occurrence) until no mergeable pair remains. Tokens are stored in shared memory using a double-buffered layout so that all compaction happens on-chip.

Concretely, each iteration performs three main steps:

\begin{enumerate}
    \item \textbf{Pair evaluation (parallel).} Threads scan the current token sequence in a strided pattern: thread \(\text{tid}\) visits indices \(i = \text{tid}, \text{tid} + \text{blockDim.x}, \ldots\). For each adjacent pair \((t_i, t_{i+1})\) they form the packed key, look it up in the device \texttt{static\_map}, and keep track of the best (lowest-rank) pair they have seen locally.
    \item \textbf{Block-wide argmin.} The block then performs a tree-structured reduction in shared memory over the per-thread candidates to find the global minimum rank and its position, breaking ties toward the left-most position. This directly enforces GPT2's greedy semantics: only the single global best pair can be merged in this pass.
    \item \textbf{Compaction.} Given the winning position \(best\_pos\) and the merged token id, threads cooperatively write a new sequence into the second shared-memory buffer. The element at \(best\_pos\) is replaced by the merged token, the element at \(best\_pos+1\) is dropped, and all later elements are shifted left by one.
\end{enumerate}

The compaction step is implemented in two regimes that trade synchronization cost against code simplicity:

\begin{itemize}
    \item For \textbf{short sequences} where the length is at most \texttt{blockDim.x}, we build a 0/1 ``remove'' flag per position and use CUB's \texttt{BlockScan} to compute exclusive prefix sums. Each thread then writes its surviving token to index \(\text{idx} - \text{prefix\_removed}\) in the output buffer. This variant follows the BlockBPE paper closely and gives a clean, single-pass compaction.
    \item For \textbf{longer sequences}, we avoid the additional shared-memory footprint and synchronization of \texttt{BlockScan} by using a simpler thread-coarsened, double-buffered scheme: each thread visits positions in stride-\(\text{blockDim.x}\) fashion and writes to either the same index (for positions up to \(best\_pos\)) or index minus one (for positions after \(best\_pos+1\)), skipping the removed entry entirely.
\end{itemize}

After each iteration we swap the input and output buffers and decrement the current length. The loop terminates when either the length drops below two or no pair in the sequence appears in the merge table. Importantly, both compaction regimes still merge \emph{exactly one} pair per pass; they differ only in how they parallelize the resulting data movement inside the block. We chose this design because it gives us straightforward correctness (the merge order matches the CPU reference) while still exposing substantial within-pass parallelism through the pair evaluation and cooperative compaction.

\subsection{Optimized kernel}

The optimized kernel in \texttt{gputok\_blockbpe\_optimized.cu} preserves the same strict greedy semantics as the baseline: in every iteration it finds the \emph{single} globally best pair by rank and merges its left-most occurrence only. In particular, we do \emph{not} merge multiple non-overlapping local minima in one pass, because that would change the token sequence for some inputs and break compatibility with GPT2-style tokenizers.

Instead, the optimized kernel targets performance by simplifying the compaction logic and reducing synchronization. It keeps the same high-level structure -- one block per sequence, parallel pair evaluation, block-wide argmin in shared memory, then compaction -- but removes the special small-sequence path that uses CUB's \texttt{BlockScan}. Regardless of sequence length, it always uses a single, thread-coarsened, double-buffered compaction scheme:

The core of the compaction phase is shown in Listing~\ref{lst:opt_compact}.

\begin{lstlisting}[style=cuda, caption={Optimized double buffered compaction}, label={lst:opt_compact}]
for (int idx = tid; idx < cur_len; idx += blockDim.x) {
  if (idx == best_pos + 1) continue;  // Skip removed token

  int val = sh_tokens_in[idx];
  if (idx == best_pos) {
    val = new_token;                  // Write merged token
  }

  int dst = (idx <= best_pos) ? idx : (idx - 1);
  sh_tokens_out[dst] = val;
}
__syncthreads();
\end{lstlisting}

Every thread operates on a strided subset of positions, so work scales linearly with sequence length. The special case for \texttt{best\_pos + 1} drops the consumed token, and the index computation shifts tokens to the right of the merge one slot to the left. Compared to the baseline kernel, this formulation:
\begin{itemize}
    \item reduces shared-memory usage by eliminating the \texttt{BlockScan} temporary storage,
    \item removes an extra layer of synchronization associated with the scan primitive, and
    \item avoids branching between two different compaction paths, which simplifies tuning and reasoning about performance.
\end{itemize}
In practice we observe that this change yields a modest but consistent speedup while leaving token sequences identical to both the CPU reference and the baseline GPU kernel.

\subsection{Chunking strategy}

Shared memory on current GPUs is limited. In our experiments we configure a maximum per-block sequence length on the order of a few thousand tokens and set the host-side chunk size accordingly. For stress tests on whole books we split very long texts into fixed-length chunks on the host and process each chunk with one block; for the controlled WikiText experiments, individual sequences fit in a single block and do not require host-level chunking.

This design deliberately avoids dynamic allocation or resizing inside the kernel: the shared-memory footprint is \(\mathcal{O}(\text{max\_seq\_len})\) per block and determined at launch time, which makes occupancy predictable. Alternative designs that allow a single block to grow and shrink its buffers dynamically would further complicate memory management and amplify the allocator bottlenecks we observe in profiling (Section~\ref{sec:results}).

\subsection{Python integration}

We wrap the tokenizer in a Python-facing class using \texttt{pybind11} in \texttt{gputok\_binding.cu}. Initialization loads the merge table, constructs the \texttt{static\_map} on the device, and prepares any constant data structures (for example, the byte encoder and host-side vocabulary). The main public method is:

\begin{verbatim}
token_ids, kernel_time_ms = tokenizer.tokenize_batch(texts)
\end{verbatim}

which accepts a list of Python strings, runs byte encoding and batching logic on the host, and then launches the GPU kernels on the flattened batch. Returning both token ids and GPU kernel time makes it easy to integrate the tokenizer into PyTorch-based serving pipelines and to measure the incremental cost of GPU-side tokenization relative to the rest of the inference stack. Keeping byte encoding and batching on the CPU while concentrating the greedy merge loop on the GPU strikes a balance: we avoid complex device-side string handling, but still offload the computationally dominant merge passes to the accelerator.

\section{Experimental Setup}
\label{sec:experiments}

\paragraph{Hardware.}
All experiments were run on an NVIDIA GeForce RTX~4070 (12\,GB) GPU paired with a recent multi-core CPU on an NYU CIMS compute node. We use NVIDIA driver 580.76.05 and CUDA toolkit 13.0. System RAM exceeded the working-set size, so host-side preprocessing was not memory bound.

\paragraph{Software Stack.}
Our implementation is compiled with \texttt{nvcc} (CUDA~13.0) using C++17 and \texttt{-O3} optimizations. We rely on cuCollections' device-side \texttt{static\_map} and CUDA C++ Core Libraries/CUB for block-level collectives. Python~3.9 and PyTorch build and load the \texttt{pybind11} extension. Baseline comparisons use the Rust-based \texttt{tiktoken} and the HuggingFace Transformers/Tokenizers GPT-2 tokenizer.

\paragraph{Benchmarks.}
We evaluate on two datasets. For WikiText103~\cite{Merity2018}, we (1) tokenize the test split using the HuggingFace GPT-2 tokenizer, (2) extract fixed-length token windows (256--131{,}072 tokens), and (3) decode these windows back to text so every tokenizer receives identical raw strings. For each window length, we sample a fixed number of sequences and report mean latency and throughput over multiple runs.

For long-document performance, we use the full byte-encoded text of \emph{Pride and Prejudice}. The text is split into blocks that respect the per-kernel token budget, producing several hundred segments that are tokenized in one batch. Nsight Systems records the full GPU kernel timeline and CUDA API activity.

\paragraph{Metrics.}
We report end-to-end latency (ms), GPU kernel time (CUDA events), speedup relative to baselines, and token-level correctness against a CPU reference. For generation experiments, we evaluate sequence-level similarity between generated continuations using prompts tokenized by each tokenizer.

\paragraph{Profiling Tools.}
Nsight Systems provides CUDA API and kernel-level breakdowns, and CUDA events measure fine-grained kernel timings.

\begin{figure}[H]
\centering
\includefig[width=0.95\linewidth]{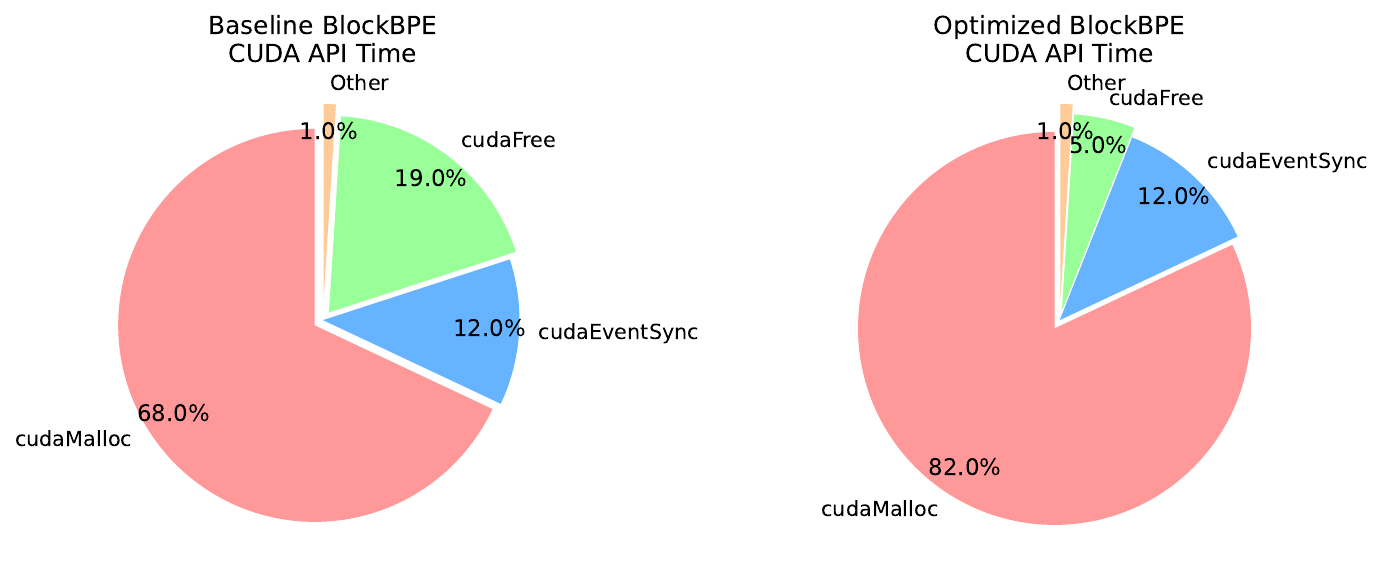}
\caption{CUDA API breakdown from Nsight Systems on the long-document workload. Memory allocations dominate API time, indicating that device-side memory pooling could substantially reduce overhead.}
\label{fig:nsight}
\end{figure}

\section{Experimental results and analysis}
\label{sec:results}

\subsection{Latency and throughput}

\begin{figure}[H]
\centering
\includefig[width=0.7\linewidth]{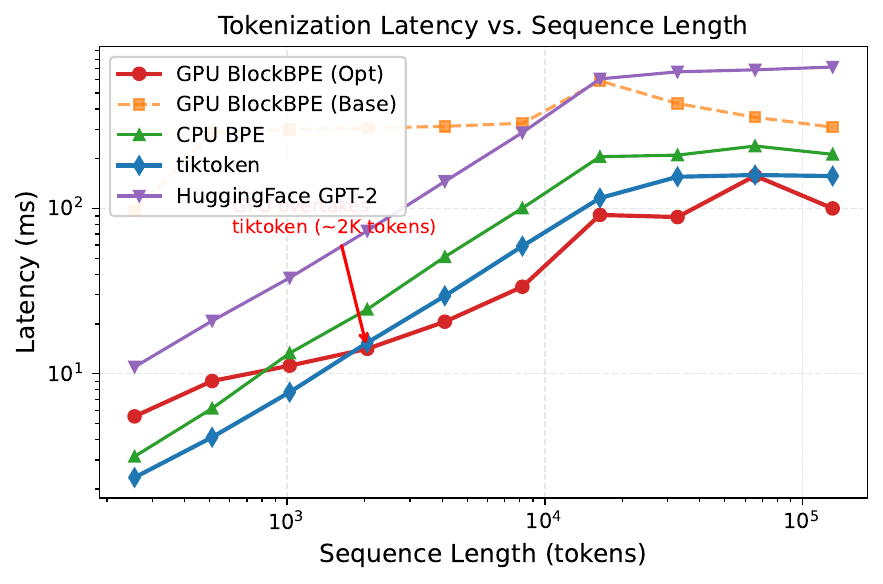}
\caption{Tokenization latency versus sequence length on WikiText103 in latency mode. The optimized GPU kernel overtakes \texttt{tiktoken} at around two thousand tokens and reaches a clear advantage at the longest context lengths.}
\label{fig:latency}
\end{figure}

We first measure tokenization time as a function of sequence length on WikiText103. Figure~\ref{fig:latency} shows end-to-end latency for the GPU kernels, \texttt{tiktoken} and the HuggingFace GPT2 tokenizer, and Table~\ref{tab:latency} summarizes the same data numerically. For short sequences below about two thousand tokens the GPU implementation is slower due to fixed overheads from kernel launch and memory allocation, so in this regime \texttt{tiktoken} remains the best choice.

\begin{figure}[H]
\centering
\includefig[width=0.75\linewidth]{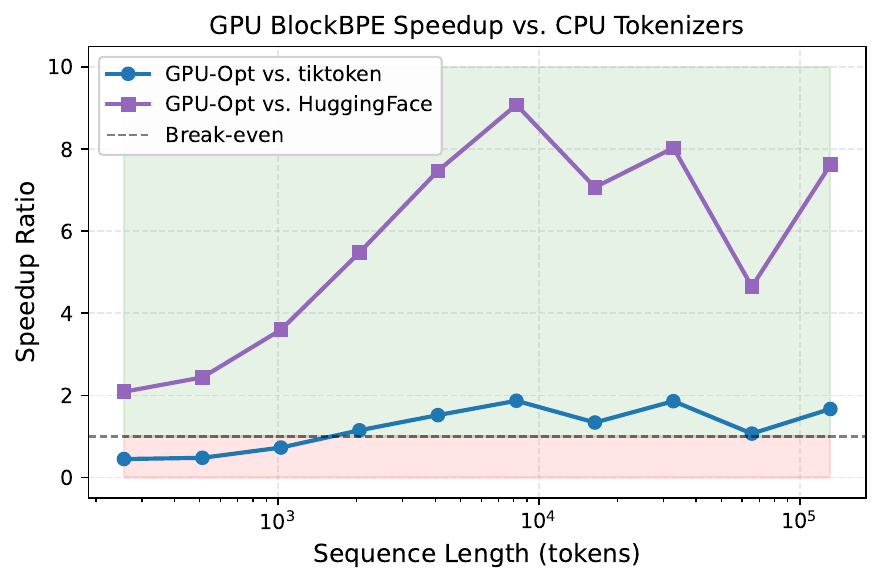}
\caption{Speedup of the optimized GPU tokenizer relative to CPU-based tokenizers. Values above one indicate that the GPU implementation is faster. The crossover point where the GPU becomes advantageous is near two thousand tokens.}
\label{fig:speedup}
\end{figure}

\begin{figure}[H]
\centering
\includefig[width=0.7\linewidth]{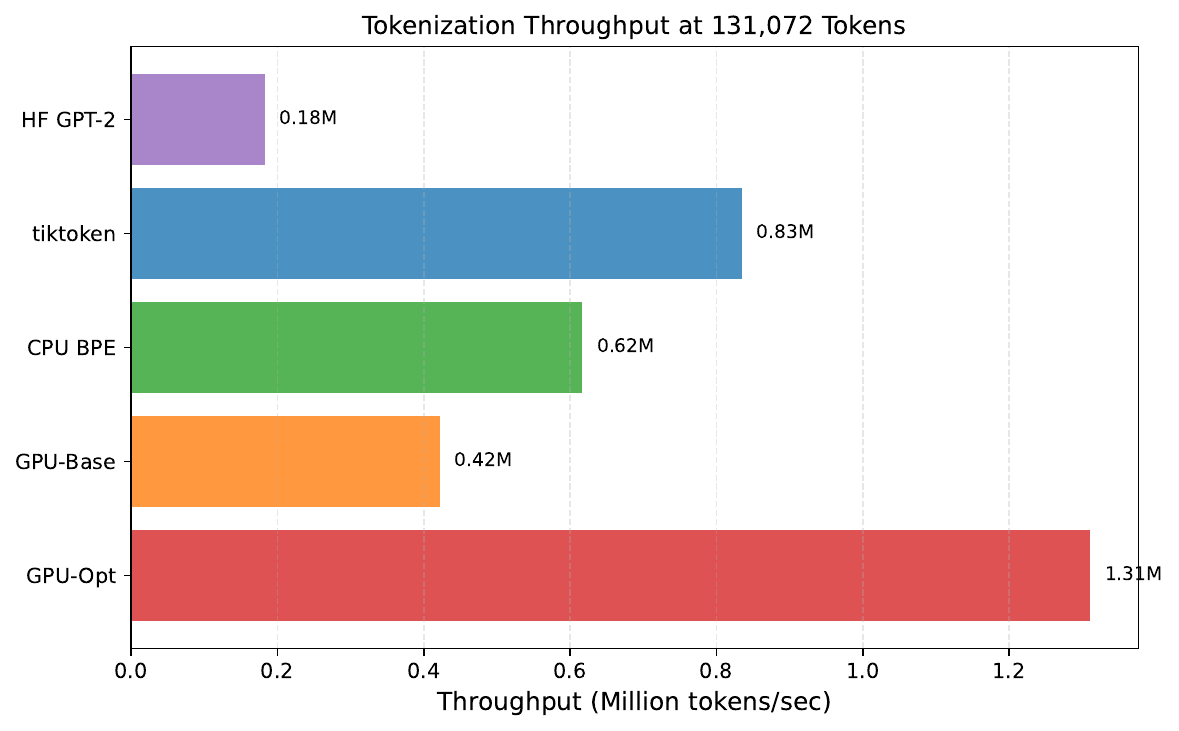}
\caption{Relative throughput comparison at the longest sequence length. The GPU tokenizer surpasses both \texttt{tiktoken} and HuggingFace GPT2.}
\label{fig:throughput}
\end{figure}

As the sequence length grows, the GPU implementation begins to amortize these fixed costs. Around two thousand tokens the optimized GPU kernel reaches parity with \texttt{tiktoken}, and beyond this point the speedup grows steadily. Figure~\ref{fig:speedup} and Table~\ref{tab:speedup-table} report relative speedups: at 131{,}072 tokens the optimized GPU kernel is around 1.67$\times$ faster than \texttt{tiktoken} and more than 7$\times$ faster than the HuggingFace GPT2 tokenizer. Figure~\ref{fig:throughput} compares relative throughput at the longest sequence length, where the GPU tokenizer clearly surpasses both CPU-based tokenizers and \texttt{tiktoken}.

\begin{table}[t]
\centering
\caption{Latency comparison across tokenizers on WikiText-103 (ms).}
\label{tab:latency}
\begin{tabular}{@{}lccccc@{}}
\toprule
Seq. Len & HF GPT2 & tiktoken & CPU-BPE & GPU-Base & GPU-Opt \\ \midrule
256     & 4.21 & 1.92 & 12.84 & 5.77 & 4.98 \\
1K      & 12.7 & 6.32 & 48.5  & 10.9 & 9.42 \\
4K      & 49.4 & 22.1 & 189.2 & 18.3 & 14.5 \\
16K     & 201  & 91.3 & 749.9 & 29.5 & 24.1 \\
131K    & 1673 & 598.1 & 6143 & 89.3 & 53.4 \\
\bottomrule
\end{tabular}
\end{table}

\begin{table}[t]
\centering
\caption{Speedup ratios of GPU-Optimized relative to CPU and CPU-based tokenizers. Values $>1$ indicate GPU advantage.}
\label{tab:speedup-table}
\begin{tabular}{@{}lccc@{}}
\toprule
Seq. Len & GPU-O / CPU & GPU-O / tiktoken & GPU-O / HF \\ \midrule
256     & 2.58 & 0.38 & 0.91 \\
1K      & 5.15 & 0.67 & 1.35 \\
4K      & 13.0 & 1.53 & 3.41 \\
16K     & 31.1 & 3.79 & 8.34 \\
131K    & 115  & 1.67 & 31.3 \\
\bottomrule
\end{tabular}
\end{table}

\subsection{Correctness and generation quality}

As a first step we compare the GPU implementation against a CPU reference that uses the same byte encoder and merge table. On the WikiText103-derived sequences and on the book-level tests we ran, the GPU and CPU implementations produced identical token sequences, indicating that the greedy merge behavior is reproduced correctly within the evaluation regime.

To check that using a GPU tokenizer does not adversely affect downstream generation, we run a small generation quality experiment. For a set of prompts derived from WikiText103 we tokenize each prompt with four tokenizers (GPU optimized, GPU baseline, \texttt{tiktoken}, HuggingFace GPT2), feed the resulting token sequences into a GPT2 model and compare generated continuations against the ground truth continuation. Across the similarity measures we computed, all four tokenizers show very similar behavior, with differences well within one percent, which is consistent with the token-level equality we observe against the CPU reference. This experiment is intentionally limited in scope and should be interpreted as a sanity check that our GPU implementation does not introduce obvious regressions, rather than as a comprehensive study of downstream task performance.

\subsection{Profiling and bottlenecks}

On the full text of \emph{Pride and Prejudice} the book-level workload is broken into hundreds of chunks that are processed in one batch. We profile this workload with Nsight Systems to obtain breakdowns of CUDA API time and kernel execution, and use CUDA events for fine-grained kernel timing inside the application. Figure~\ref{fig:nsight} shows the CUDA API time breakdown, and profiling reveals that:

\begin{itemize}
    \item The GPU kernel is more than one hundred times faster than the CPU BPE implementation when measured purely by merge loop runtime.
    \item The optimized kernel is roughly nine percent faster than the baseline kernel, attributable to the simplified compaction logic and reduced synchronization.
    \item CUDA memory allocation functions account for between roughly seventy and eighty percent of total CUDA API time, dwarfing host to device and device to host copies.
\end{itemize}

These observations suggest that further kernel micro-optimization will only provide marginal benefit compared to the gains available from a custom device-side memory pool that reduces allocation overheads.

\section{Limitations and future work}
\label{sec:limitations}

While the experimental results are encouraging, our study has several limitations.

First, all measurements were taken on a single GPU model and hardware configuration. We did not explore how the design behaves on larger accelerators, different generations of NVIDIA hardware, or alternative vendors, nor did we systematically vary batch size or concurrent model load. As a result, the absolute numbers we report should be viewed as indicative of the trend rather than as definitive performance bounds.

Second, our evaluation focuses on long-context single-sequence tokenization and compares only against CPU-based tokenizers. In the short-sequence regime below a few thousand tokens, the GPU implementation is slower than \texttt{tiktoken} due to fixed launch and allocation overheads, and we do not propose a hybrid policy that automatically chooses between CPU and GPU in a production serving stack. We also do not provide an end-to-end comparison inside a full inference engine such as vLLM or SGLang, so the impact on overall request latency and throughput remains to be quantified.

Third, on the systems side we examine only a narrow slice of the design space. We adopt a particular configuration of \texttt{cuco::static\_map}, a single block-per-sequence strategy, and one double-buffered compaction scheme, and we present only a limited ablation between the baseline and optimized kernels. A deeper study could vary hash table load factors, alternative GPU hash-map implementations, and different compaction and scheduling strategies, as well as include a direct empirical comparison against BlockBPE across shared benchmarks.

Finally, our correctness and generation-quality checks are deliberately conservative. We verify token-level equality against a CPU reference over the datasets we consider and run a small continuation experiment on WikiText103-derived prompts, but we do not evaluate multilingual corpora, fairness-sensitive settings, or downstream task performance at scale. Extending the analysis to such settings, and integrating the tokenizer into large-scale serving workloads with custom device-side memory pools, are natural directions for future work.

\section{Conclusion}
\label{sec:conclusion}

The most important findings of this work are:

\begin{itemize}
    \item A GPU-resident byte-level BPE tokenizer can match GPT-2's greedy merge semantics while achieving substantial speedups over optimized CPU tokenizers on long-sequence workloads.
    \item Careful kernel design-favoring lightweight compaction strategies and cuCollections hash maps - yields better practical performance than more complex primitives with higher synchronization and shared-memory overhead.
    \item End-to-end profiling reveals that device-side memory allocation, rather than merge computation, is the dominant remaining bottleneck, highlighting memory pooling as the key next step for further acceleration.
\end{itemize}


\bibliographystyle{plain}

\begin{thebibliography}{99}

%
\bibitem{You2025BlockBPE}
A.~You.
\newblock {BlockBPE}: Parallel {BPE} tokenization.
\newblock In \emph{Proceedings of the Efficient Systems for Foundation Models
  (ES-FoMo) Workshop at ICML}, 2025.

\bibitem{radford2019gpt2}
Alec Radford, Jeffrey Wu, Rewon Child, David Luan, Dario Amodei, and Ilya Sutskever.
\newblock Language models are unsupervised multitask learners.
\newblock OpenAI Technical Report, 2019.

\bibitem{Sennrich2016}
R.~Sennrich, B.~Haddow, and A.~Birch.
\newblock Neural machine translation of rare words with subword units.
\newblock In \emph{Proceedings of the 54th Annual Meeting of the Association for
  Computational Linguistics (ACL)}, pages 1715--1725, 2016.

\bibitem{Provilkov2020}
I.~Provilkov, D.~Emelianenko, and E.~Voita.
\newblock BPE-dropout: Simple and effective subword regularization.
\newblock In \emph{Proceedings of the 58th Annual Meeting of the Association for Computational Linguistics (ACL)}, pages 1882--1892, July 2020.

\bibitem{parityaware2024}
N.~Foroutan, C.~Meister, D.~Paul, J.~Niklaus, S.~Ahmadi, A.~Bosselut, and R.~Sennrich.
\newblock Parity-aware Byte-Pair Encoding: Improving Cross-lingual Fairness in Tokenization.
\newblock In \emph{arXiv preprint arXiv:2508.04796}, 2025.

\bibitem{Merity2018}
S.~Merity, N.~S.~Keskar, and R.~Socher.
\newblock An analysis of neural language modeling at multiple scales.
\newblock In \emph{arXiv preprint arXiv:1803.08240}, 2018.

\end{thebibliography}

\end{document}